\documentclass[runningheads]{llncs}
\usepackage{graphicx}
\usepackage{amsmath,amssymb} % define this before the line numbering.
\usepackage{color}
\usepackage{subcaption}
\usepackage{comment}
\usepackage{tablefootnote}
\usepackage{array,multirow}
\graphicspath{ {./fg/} }
\newcommand{\etal}{\textit{et al.}}

\begin{document}
\pagestyle{headings}
\mainmatter

\def\ACCV20SubNumber{***}  % Insert your submission number here

%===========================================================
\title{Image Captioning through Image Transformer} % Replace with your title
\titlerunning{Image Captioning through Image Transformer}
\makeatletter
\newcommand{\printfnsymbol}[1]{%
  \textsuperscript{\@fnsymbol{#1}}%
}
\makeatother

\author{Sen He\thanks{Equal contribution} \inst{1} \and
Wentong Liao\printfnsymbol{1}\inst{2} \and
Hamed R. Tavakoli\inst{3} \and Michael Yang\inst{4} \\  Bodo Rosenhahn\inst{2} \and Nicolas Pugeault\inst{5}}
\authorrunning{Sen He et al.}
\institute{CVSSP, University of Surrey, UK \and
Leibniz University of Hanover, Germany \and Nokia Technologies, Finland \and University of Twente, Netherlands \and School of Computing Science, University of Glasgow\\
\email{senhe752@gmail.com}
}

\maketitle

%===========================================================
\begin{abstract}
Automatic captioning of images is a task that combines the challenges of image analysis and text generation. One important aspect of captioning is the notion of attention: how to decide what to describe and in which order. Inspired by the successes in text analysis and translation, previous works have proposed the \textit{transformer} architecture for image captioning. However, the structure between the \textit{semantic units} in images (usually the detected regions from object detection model) and sentences (each single word) is different. Limited work has been done to adapt the transformer's internal architecture to images. In this work, we introduce the \textbf{\textit{image transformer}}, which consists of a modified encoding transformer and an implicit decoding transformer, motivated by the relative spatial relationship between image regions. Our design widens the original transformer layer's inner architecture to adapt to the structure of images. With only regions feature as inputs, our model achieves new state-of-the-art performance on both MSCOCO offline and online testing benchmarks. The code is available at  \url{https://github.com/wtliao/ImageTransformer}.
\end{abstract}

%===========================================================
\section{Introduction}\label{intro}

%Recently, there has been a significant interest in vision and language tasks such as image captioning, visual question answering, and multimodal retrieval. In this paper, we focus on the 
%Most image captioning pipelines are based on generative models, consisting of a visual encoder and a language decoder.
%Many of such pipelines rely on some attention mechanism, e.g.,\cite{xu2015show,chen2017sca,anderson2018bottom,lu2018neural}. 

Image captioning is the task of describing the content of an image in words. The problem of automatic image captioning by AI systems has received a lot of attention in the recent years, due to the success of deep learning models for both language and image processing. 
%Image captioning, which connected computer vision and natural language processing, has gained a lot of research interest recently. 
Most image captioning approaches in the literature are based on a \textit{translational} approach, with a visual encoder and a linguistic decoder. One challenge in automatic translation is that it cannot be done word by word, but that other words influence then meaning, and therefore the translation, of a word; this is even more true when translating across modalities, from images to text, where the system must decide \textit{what} must be described in the image. 
A common solution to this challenge relies on attention mechanisms. For example, previous image captioning models try to solve \textit{where} to look in the image \cite{xu2015show,chen2017sca,anderson2018bottom,lu2018neural} (now partly solved by the Faster-RCNN object detection model~\cite{ren2015faster}) in the encoding stage and use a recurrent neural network with attention mechanism in the decoding stage to generate the caption. 
But more than just to decide what to describe in the image, recent image captioning models propose to use attention 
to learn how regions of the image relate to each other, effectively encoding their \textit{context} in the image. 
%Beyond where to look in the image in the encoding part, recent attention mechanism in image captioning models focus on how to relate each the informative region with other related regions in the image through attention mechanism. 
Graph convolutional neural networks \cite{kipf2016semi} were first introduced to relate regions in the image; however, those approaches \cite{yao2018exploring,yang2019auto,guo2019aligning,yao2019hierarchy} usually require auxiliary models (e.g. visual relationship detection and/or attribute detection models) to build the visual scene graph in the image in the first place. 
In contrast, in the natural language processing field, the transformer architecture \cite{vaswani2017attention} was developed to relate embedded words in sentences, and can be trained end to end without auxiliary models explicitly detecting such relations. Recent image captioning models \cite{huang2019attention,li2019entangled,herdade2019image} adopted the transformer architectures to implicitly relate informative regions in the image through dot-product attention achieving state-of-the-art performance.

However, the transformer architecture was designed for machine translation of text. In a text, a word is either to the left or to the right of another word, with different distances. In contrast, images are two-dimensional (indeed, represent three-dimensional scenes), so that a region may not only be on the left or right of another region, it may also contain or be contained in another region. 
The relative spatial relationship between the semantic units in images has a larger degree of freedom than that in sentences. Furthermore, in the decoding stage of machine translation, a word is usually translated into another word in other languages (one to one decoding), whereas for an image region, we may describe its context, its attribute and/or its relationship with other regions (one to more decoding). 
One limitation of previous transformer-based image captioning models \cite{huang2019attention,li2019entangled,herdade2019image} is that they adopt the transformer's internal architecture designed for the machine translation, where each transformer layer contains a single (multi-head) dot-product attention module. 
In this paper, we introduce the \textbf{\textit{image transformer}} for image captioning, where each transformer layer implements multiple sub-transformers, to encode spatial relationships between image regions and decode the diverse information in image regions.

The difference between our method and previous transformer based models \cite{huang2019attention,herdade2019image,li2019entangled} is that our method focuses on the \textit{inner architectures} of the transformer layer, in which we widen the transformer module. Yao~\etal~\cite{yao2019hierarchy} used a hierarchical concept in the encoding part of their model, our model focuses on the local spatial relationships for each query region whereas their method is a global tree hierarchy. Furthermore, our model does not require auxiliary models (\textit{ie}, for visual relation detection and instance segmentation) to build the visual scene graph. Our encoding method can be viewed as the combination of a visual semantic graph and a spatial graph which use a transformer layer to implicitly combine them without auxiliary relationship and attribute detectors.

\begin{figure}[t!]
\centering
\includegraphics[width=\textwidth]{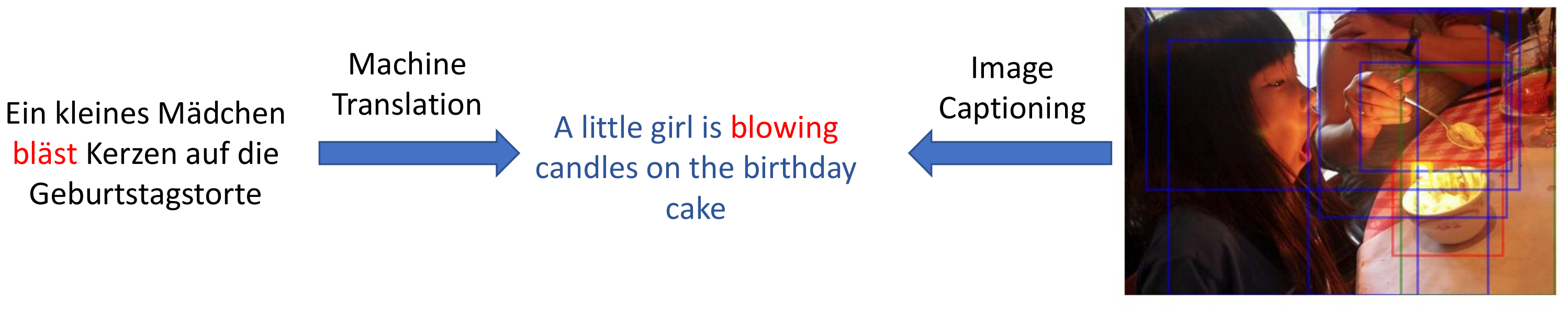}
\vspace{-15pt}
\caption{Image captioning vs machine translation.}
\vspace{-10pt}
\label{fig:data}
\end{figure}

The contributions of this paper can be summarised as follows:
\begin{itemize}
\item
    We propose a novel internal architecture for the transformer layer adapted to the image captioning task, with a modified attention module suited to the complex natural structure of image regions.
\item 
    We report thorough experiments and ablation study were done in the work to validate our proposed architecture, state-of-the-art performance was achieved on the MSCOCO image captioning offline and online testing dataset with only region features as input.
\end{itemize}

The rest of the paper is organized as follows: Sec.~\ref{sec:background} reviews the related attention-based image captioning models; Sec.~\ref{sec:transformer} introduces the standard transformer model and our proposed image transformer; followed by the experiment results and analysis in Sec.~\ref{sec:experiments}; finally, we will conclude this paper in Sec.~\ref{sec:conclusion}.

\section{Related Work}\label{sec:background}
We characterize current attention-based image captioning models into single-stage attention models, two-stages attention models, visual scene graph based models, and transformer-based models. We will review them one by one in this section.

\subsection{Single-Stage Attention Based Image Captioning}
Single-stage attention-based image captioning models are the models where attention is applied at the decoding stage, where the decoder attends to the most informative region \cite{luo2016understanding} in the image when generating a corresponding word.

The availability of large-scale annotated datasets \cite{deng2009imagenet,chen2015microsoft} enabled the training of deep models for image captioning. Vinyals~\etal~\cite{vinyals2015show} proposed the first deep model for image captioning. Their model uses a CNN pre-trained on ImageNet~\cite{deng2009imagenet} to encode the image, then a LSTM~\cite{gers1999learning} based language model is used to decode the image features into a sequence of words. Xu~\etal~\cite{xu2015show} introduced an attention mechanism into image captioning during the generation of each word, based on the hidden state of their language model and the previous generated word. Their attention module generates a matrix to weight each receptive field in the encoded feature map, and then feed the weighted feature map and the previous generated word to the language model to generate the next word. Instead of only attending to the receptive field in the encoded feature map, Chen~\etal~\cite{chen2017sca} added a feature channel attention module, their channel attention module re-weight each feature channel during the generation of each word. 
Not all the words in the sentence have a correspondence in the image, so Lu~\etal~\cite{lu2017knowing} proposed an adaptive attention approach, where their model has a visual sentinel which adaptively decides when and where to rely on the visual information. 

The single-stage attention model is computational efficient, but lacks accurate positioning of informative regions in the original image.

\subsection{Two-Stages Attention Based Image Captioning}
Two stage attention models consists of \textit{bottom-up} attention and \textit{top-down} attention, where bottom-up attention first uses object detection models to detect multiple informative regions in the image, then top-down attention attends to the most relevant detected regions when generating a word.

Instead of relying on the coarse receptive fields as informative regions in the image, as single-stage attention models do,
Anderson~\etal~\cite{anderson2018bottom} train the detection models on the \textit{Visual Genome} dataset~\cite{krishna2017visual}. The trained detection models can detect $10-100$ informative regions in the image.
They then use a two-layers LSTM network as decoder, where the first layer generates a state vector based on the embedded  word vector and the mean feature of the detected regions and the second layer uses the state vector from the previous layer to generate a weight for each detected region. The weighted sum of detected regions feature is used as a context vector for predicting the next word.
Lu~\etal~\cite{lu2018neural} developed a similar network, but with a detection model trained on \textit{MSCOCO}~\cite{lin2014microsoft}, which is a smaller dataset than \textit{Visual Genome}, and therefore less informative regions are detected. 

The performance of two-stage attention based image captioning models is improved a lot against single-stage attention based models. However, each detected region is isolated from others, lacking the relationship with other regions.

\subsection{Visual Scene Graph Based Image Captioning}
Visual scene graph based image captioning models extend two-stage attention models by injecting a graph convolutional neural network to relate detected informative regions, and therefore refine their features before feeding into the decoder.

Yao~\etal~\cite{yao2018exploring} developed a model which consists of a semantic scene graph and a spatial scene graph. In the semantic scene graph, each region is connected with other semantically related regions, those relationships are usually determined by a visual relationship detector among a union box. In the spatial scene graph, the relationship between two regions is defined by their relative positions. Then the feature of each node in the scene graph is refined with their related nodes through graph neural networks~\cite{kipf2016semi}. Yang~\etal~\cite{yang2019auto} use an auto-encoder, where they first encode the graph structure in the sentence based on the SPICE \cite{anderson2016spice} evaluation metric to learn a dictionary, then the semantic scene graph is encoded using the learnt dictionary. 
The previous two works treat the semantic relationships as edges in the scene graph, while Guo~\etal~\cite{guo2019aligning} treat them as nodes in the scene graph. Also, their decoder focuses on different aspects of a region. Yao~\etal \cite{yao2019hierarchy} further introduces the tree hierarchy and instance level feature into the scene graph. 

Introducing the graph neural network to relate informative regions yields a sizeable performance improvement for image captioning models, compared to two-stage attention models. However, it requires auxiliary models to detect and build the scene graph at first. Also those models usually have two parallel streams, one responsible for the semantic scene graph and another for spatial scene graph, which is computationally inefficient.

\subsection{Transformer Based Image Captioning}
Transformer based image captioning models use the dot-product attention mechanism to relate informative regions implicitly.

Since the introduction of original transformer model~\cite{vaswani2017attention}, more advanced architectures were proposed for machine translation based on the structure or the natural characteristic of sentences  \cite{hao2019multi,wang2019self,wang2019tree}. 
In image captioning, AoANet~\cite{huang2019attention} uses the original internal transformer layer architecture, with the addition of a \textit{gated linear layer}~\cite{dauphin2017language} on top of the multi-head attention. The object relation network~\cite{herdade2019image} injects the relative spatial attention into the dot-product attention.
Another interesting result described by Herdade~\etal~\cite{herdade2019image} is that the simple position encoding (as proposed in the original transformer) did not improve image captioning performance. The entangled transformer model~\cite{li2019entangled} features a dual parallel transformer to encode and refine visual and semantic information in the image, which is fused through gated bilateral controller.

Compared to scene graph based image captioning models, transformer based models do not require auxiliary models to detect and build the scene graph at first, which is more computational efficient.
However current transformer based models still use the inner architecture of the original transformer, designed for text, where each transformer layer has a single multi-head dot-product attention refining module. 
This structure does not allow to model the full complexity of relations between image regions, therefore we propose to change the inner architecture of the transformer layer to adapt it to image data. We widen the transformer layer, such that each transformer layer has multiple refining modules for different aspects of regions both in the encoding and decoding stages.

\section{Image Transformer}\label{sec:transformer}
In this section, we first review the original transformer layer~\cite{vaswani2017attention}, we then elaborate the encoding and decoding part for the proposed \textbf{\textit{image transformer}} architecture.

\begin{figure}[h!]
\centering
\includegraphics[width=1\textwidth]{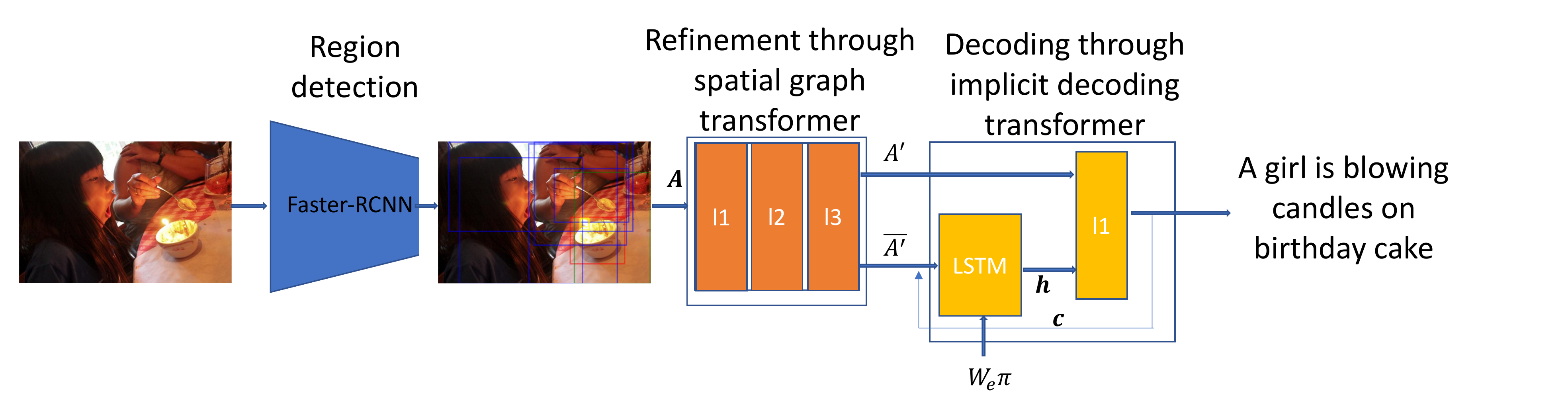}
\vspace{-10pt}
\caption{The overall architecture of our model, the refinement part consists of 3 stacks of spatial graph transformer layer, and the decoding part has a LSTM layer with a implicit decoding transformer layer.}
\vspace{-10pt}
\label{fig:model_archi}
\end{figure}

\subsection{Transformer Layer}
A transformer consists of a stack of multi-head dot-product attention based transformer refining layer.

\noindent In each layer, for a given input $A \in \mathbb{R}^{N\times D}$, consisting of $N$ entries of $D$ dimensions. In natural language processing, the input entry can be the embedded feature of a word in a sentence, and in computer vision or image captioning, the input entry can be the feature describing a region in an image. The key function of transformer is to refine each entry  with other entries through multi-head dot-product attention. Each layer of a transformer first transforms the input into queries ($Q = AW_Q$, $W_Q \in \mathbb{R}^{D\times D_k}$), keys ($K= AW_K$, $W_K \in \mathbb{R}^{D\times D_k}$) and values ($V = AW_V$, $W_A \in \mathbb{R}^{D\times D_v}$) though linear transformations, then the scaled dot-product attention is defined by:

\begin{equation}\label{eq:1}
\centering
\begin{aligned}
  \text{Attention}(Q,K,V) = \text{Softmax}\left(\frac{QK^T}{\sqrt{D_k}}\right) V,
\end{aligned}
%\label{eq:attention}
\end{equation}
% where, $d$ is a normalising factor. In stead of performing a single scaled dot product attention, the transformer adopts a multihead dot-product attention:
where $D_k$ is the dimension of the key vector and $D_v$ the dimension of the value vector ($D=D_k=D_v$ in the implementation). To improve the performance of the attention layer, multi-head attention is applied:
\begin{equation}
\centering
\begin{aligned}
& \text{MultiHead}(Q,K,V) = \text{Concat}(\text{head}_1,\dots,\text{head}_h)W_O,\\
& \text{head}_i = \text{Attention}(AW_{Q_i},AW_{K_i},AW_{V_i}).
\end{aligned}
\label{eq:multiheawd}
\end{equation}

\noindent The output from the multi-head attention is then added with the input and normalised:
\begin{equation}\label{eq:3}
    \centering
    \begin{aligned}
     A_m = \text{Norm}(A + \text{MultiHead}(Q,K,V)),
    \end{aligned}
\end{equation}
where $\text{Norm}(\cdot)$ denote layer normalisation. 

The transformer implements residual connections in each module, such that the final output of a transformer layer is:
\begin{equation}
    \centering
    \begin{aligned}
    A^{'} = \text{Norm}(A_m + \phi(A_m W_f)),
    \end{aligned}
\end{equation}
where $\phi$ is a feed-forward network with non-linearity. 
 
Each refining layer takes the output of its previous layer as input (the first layer takes the original input). The decoding part is also a stack of transformer refining layers, which take the output of encoding part as well as the embedded features of previous predicted word.

\subsection{Spatial Graph Encoding Transformer Layer}
\begin{figure}%[t!]
\centering
\begin{subfigure}[b]{0.49\linewidth}
\includegraphics[width=\linewidth]{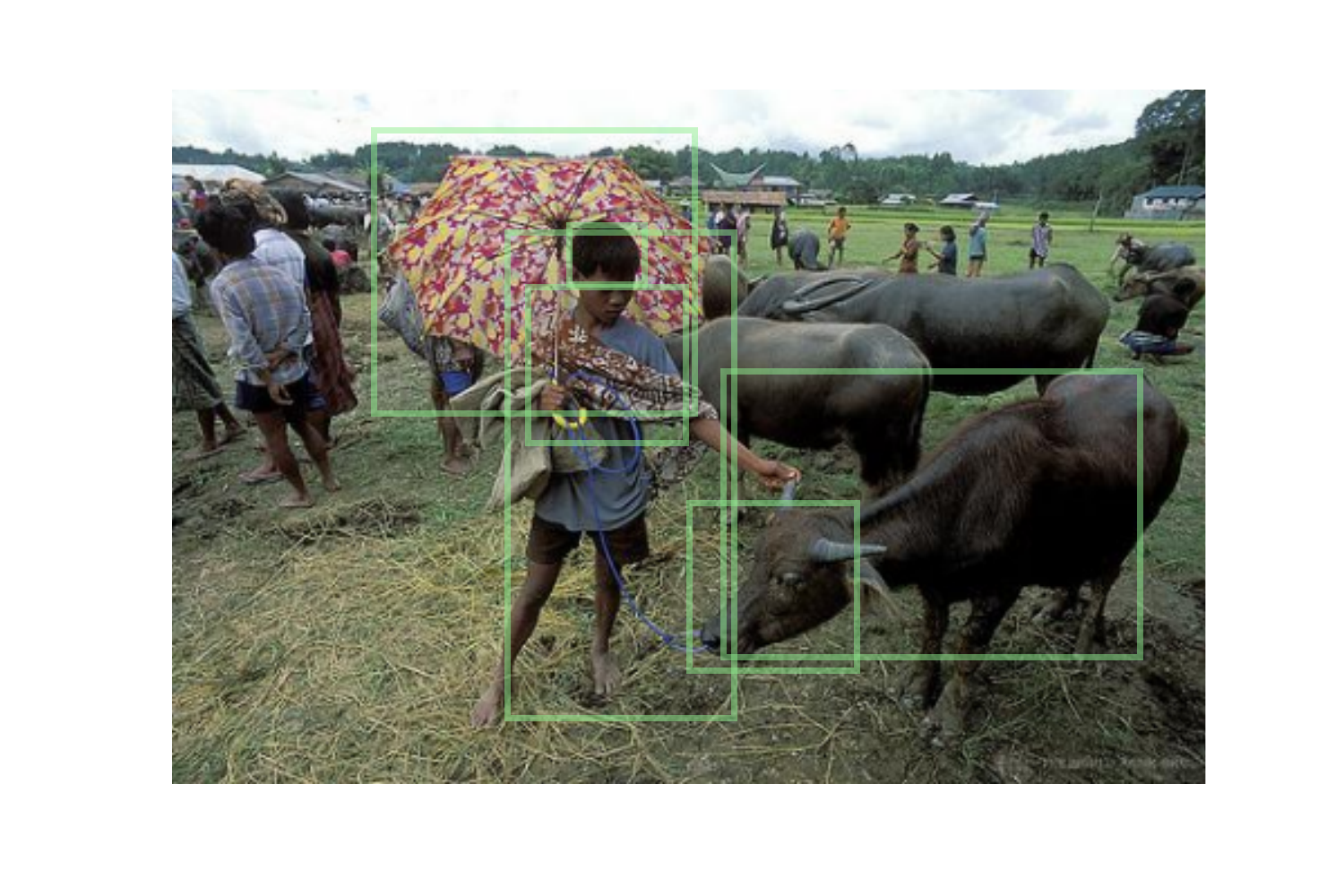}
\caption{}
%\label{fig:h_graph}
\end{subfigure}
\begin{subfigure}[b]{0.49\linewidth}
\includegraphics[width=\linewidth]{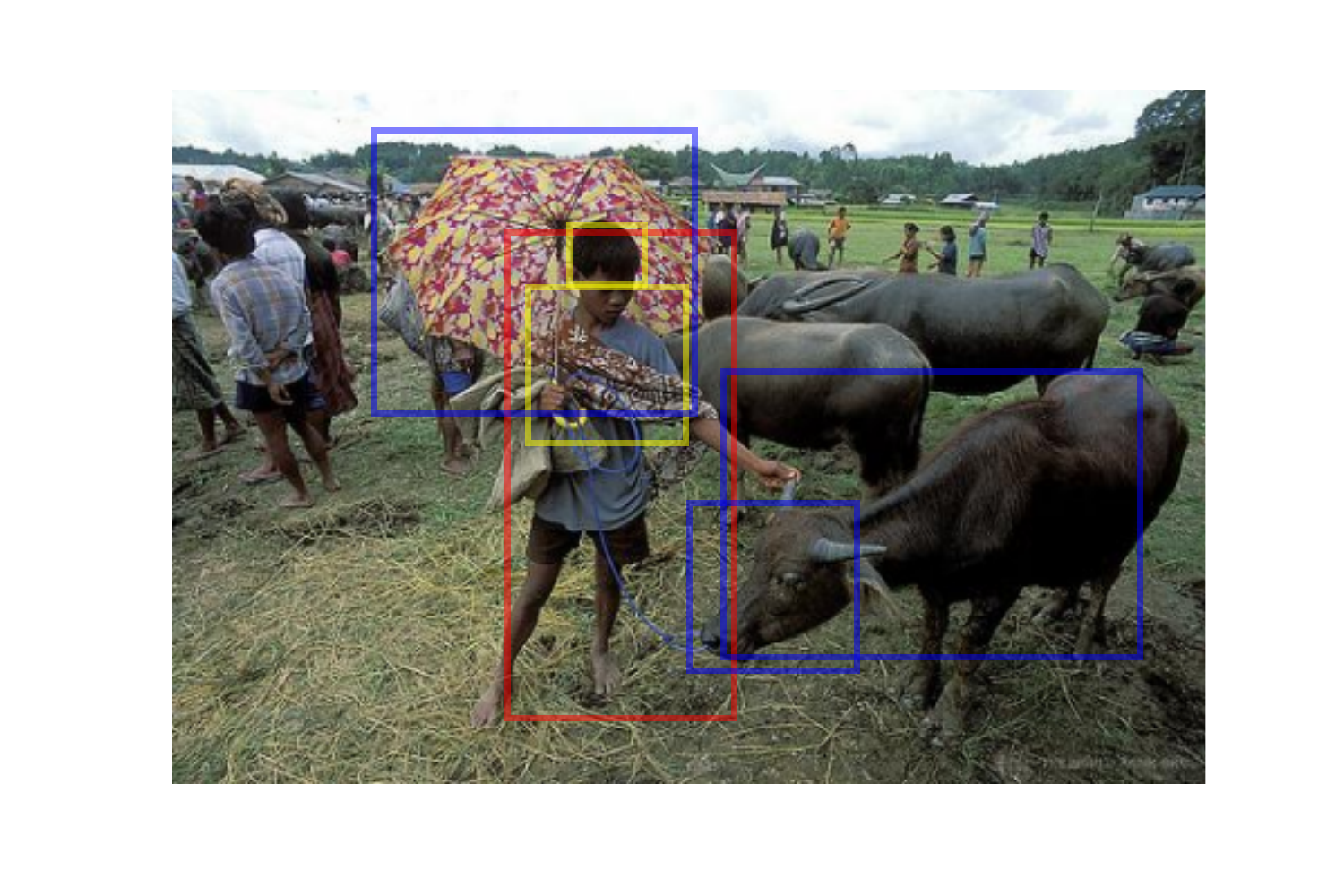}
\caption{}
%\label{fig:overlap}
\end{subfigure}
\caption{(a) Image with detected regions; (b) An example of query region in the image (man in the red bounding box), and its neighbor regions (regions in blue bounding boxes, bull, umbrella, etc), child regions (regions in the yellow bounding boxes, hair,cloth).}
\label{fig:spatial_rela_exa}
\end{figure}

In contrast to the original transformer, which only considers spatial relationships between query and key pairs as \textbf{\textit{neighborhood}}, we propose to use a spatial graph transformer in the encoding part, where we consider three common categories of spatial relationship for each query region in a graph structure: \textit{parent}, \textit{neighbor}, and \textit{child} (an example shown in Fig.~\ref{fig:spatial_rela_exa}). Thus we widen each transformer layer by adding three sub-transformer layers in parallel in each layer, each sub-transformer responsible for a category of spatial relationship, all sharing the same query. 
In the encoding stage, we define the relative spatial relationship between two regions based on their overlap. We first compute the graph adjacent matrices $\Omega_p \in \mathbb{R}^{N\times N}$ (parent node adjacent matrix), $\Omega_n \in \mathbb{R}^{\in N \times N}$ (neighbor node adjacent matrix), and $\Omega_c \in \mathbb{R}^{\in N \times N}$ (child node adjacent matrix) for all regions in the image:

\newcommand{\area}{\text{Area}}

\begin{equation}
    \centering
    \begin{aligned}
    \Omega_{p}[l,m] &=  
    \left\{ \begin{aligned}
    1, &\ \text{if}\ \frac{\area(l \cap m)}{\area(l)}\geqslant \epsilon \ \text{and} \ \frac{\area(l \cap m)}{\area(l)}> \frac{\area(l \cap m)}{\area(m)}\\
    0, &\ \text{otherwise.}
    \end{aligned}
    \right.\\
    \Omega_{c}[l,m] &= \Omega_{p}[m,l] \\
    %& \Omega_{c}[l,m] = 1,\ \text{if}\ \frac{\area(l \cap m)}{\area(m)}\geqslant \epsilon \ \text{and} \ \frac{\area(l \cap m)}{\area(m)}> \frac{\area(l \cap m)}{\area(l)},\ \text{else} \ 0\\
    \text{with} & \sum_{i \in \{p,n,c\}}\Omega_{i}[l,m] = 1 
    \end{aligned}
\end{equation}
\noindent where $\epsilon=0.9$ in our experiment. 
The spatial graph adjacent matrices are used as the spatial hard attention embedded into each sub-transformer to combine the output of each sub-transformer in the encoder. More specifically, the original encoding transformer defined in Eqs.~\eqref{eq:1} and~\eqref{eq:multiheawd} are reformulated as:
% in the image encoding transformer is defined as :
\begin{equation}\label{eq:6}
    \centering
    \begin{aligned}
    & \text{Attention}(Q,K_i,V_i) = \Omega_i \circ \text{Softmax}\left(\frac{QK_{i}^{T}}{\sqrt{d}}\right)V_{i}, %i \in \{p,n,c\}\\
    \end{aligned}
\end{equation}
$\circ$ is the Hadamard product, and
\begin{equation}\label{eq:7}
    \centering
    \begin{aligned}
    A_m = \text{Norm}\left(A+\sum_{i \in \{p,n,c\}}\text{MultiHead}(Q,K_i,V_i)\right).
    \end{aligned}
\end{equation}
As we widen the transformer, we halve the number of stacks in the encoder to achieve similar complexity as the original one (3 stacks, while the original transformer features 6 stacks). With our formulation, we combined  the spatial graph and semantic graph (the scene graph based methods \cite{yao2018exploring,guo2019aligning} require two branches to encode them) into a transformer layer. Note that the original transformer architecture is a special case of the proposed architecture, when no region in the image either contains or is contained by another.

\begin{figure}[t!]
\centering
\includegraphics[width=1\textwidth]{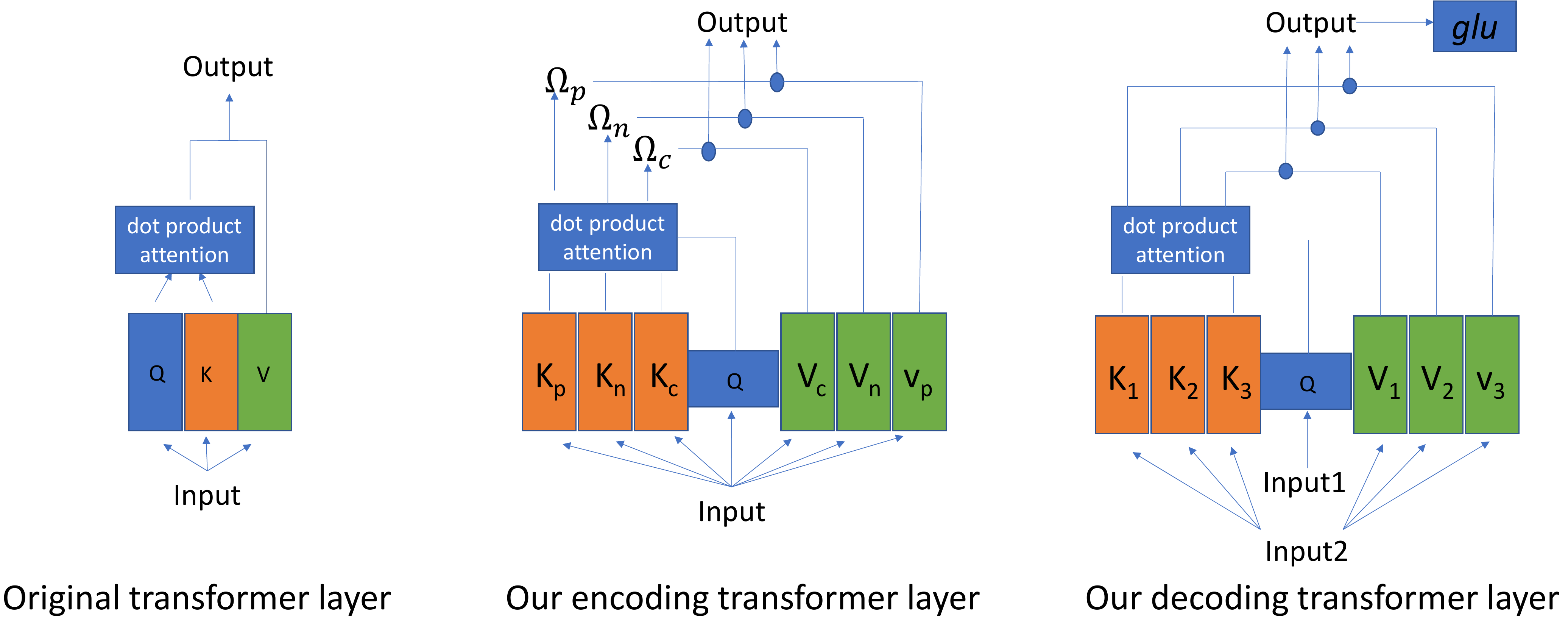}
\vspace{-15pt}
\caption{The difference between the original transformer layer and the proposed encoding and decoding transformer layers.}
\vspace{-10pt}
\label{fig:diff}
\end{figure}
\subsection{Implicit Decoding Transformer Layer}
Our decoder consists of a LSTM \cite{hochreiter1997long} layer and an implicit transformer decoding layer, which we proposed to decode the diverse information in a region in the image. The LSTM layer is a common memory module and the transformer layer infers the most relevant region in the image through dot product attention.

At first, the LSTM layer receives the  mean of the output ($\overline{A}=\frac{1}{N}\sum_{i=1}^{N}A_{i}^{'}$) from the encoding transformer, a context vector ($c_{t-1}$) at last time step and the embedded feature vector of current word in the ground truth sentence:
\begin{equation}
    \centering
    \begin{aligned}
    & x_t = [W_{e}\pi_{t},\overline{A}+c_{t-1}]\\
    & h_{t},m_{t} = \text{LSTM}(x_{t},h_{t-1},m_{t-1})\\
    \end{aligned}
\end{equation}
Where, $W_{e}$ is the word embedding matrix, $\pi_t$ is the $t^\text{th}$ word in the ground truth. 
The output state $h_t$ is then transformed linearly and treated as the query for the input of the implicit decoding transformer layer. The difference between the original transformer layer and our implicit decoding transformer layer is that we also widen the decoding transformer layer by adding several sub-transformers in parallel in one layer, such that each sub-transformer can implicitly decode different aspects of a region. 
It is formalised as follows:

\begin{equation}
    \centering
    \begin{aligned}
     A^D_{t,i} = \text{MultiHead}(W_{DQ}h_{t},W_{DKi}A^{'},W_{DVi}A^{'})
    \end{aligned}
\end{equation}
Then, the mean of the sub-transformers' output is passed through a gated linear layer (GLU) \cite{dauphin2017language} to extract the new context vector ($c_{t}$) at the current step by channel:
\begin{equation}\label{eq:eq10}
    \centering
    \begin{aligned}
    c_{t} = \text{GLU}\left(h_{t},\frac{1}{M}\sum_{i=1}^{M}A^D_{t,i}\right)
    \end{aligned}
\end{equation}
The context vector is then used to predict the probability of word at time step $t$:
\begin{equation}
    \centering
    \begin{aligned}
    p(y_{t}|y_{1:t-1}) = \text{Softmax}(w_{p}c_{t}+b_{p})
    \end{aligned}
\end{equation}

The overall architecture of our model is illustrated in Fig.~\ref{fig:model_archi}, and the difference between the original transformer layer and our proposed encoding and decoding transformer layer is showed in Fig.~\ref{fig:diff}.

\subsection{Training Objectives}
Given a target ground truth as a sequence of words $y_{1:T}^{\ast}$, for training the model parameters $\theta$, we follow the previous method, such that we first train the model with cross-entropy loss:
\begin{equation}
    \centering
    \begin{aligned}
    L_{XE}(\theta) = -\sum_{t=1}^{T}\log(p_{\theta}(y_{t}^{\ast}|y_{1:t-1}^{\ast}))
    \end{aligned}
\end{equation}
then followed by self-critical reinforced training \cite{rennie2017self} optimizing the CIDEr score \cite{vedantam2015cider}:
\begin{equation}
    \centering
    \begin{aligned}
    L_{R}(\theta) = -E_{(y_{1:T}\sim p_{\theta})}[r(y_{1:T})]
    \end{aligned}
\end{equation}
where $r$ is the score function and the gradient is approximated by:
\begin{equation}
    \centering
    \begin{aligned}
    \bigtriangledown_{\theta}\approx -(r(y_{1:T}^{s})-(\hat{y}_{1:T}))\bigtriangledown_{\theta}\log p_{\theta}(y_{1:T}^{s})
    \end{aligned}
\end{equation}

\begin{table}[t!]
    \centering
    \begin{tabular}{c|c c c c c c}
    \hline
    model & Bleu1 & Bleu4 & METEOR & ROUGE-L & CIDEr & SPICE\\
    \hline
    \multicolumn{7}{c}{\textbf{single-stage model}}\\
    \hline
    Att2all\cite{rennie2017self}&-  &34.2 & 26.7 & 55.7 & 114.0& -\\
    \hline
    \multicolumn{7}{c}{\textbf{two-stages model}}\\
    \hline
    n-babytalk\cite{lu2018neural}& 75.5 & 34.7 & 27.1 & - & 107.2& 20.1\\
    up-down\cite{anderson2018bottom}&79.8 & 36.3 & 27.7 & 56.9 & 120.1& 21.4\\
    \hline
    \multicolumn{7}{c}{\textbf{scene graph based model}}\\
    \hline
    GCN-LSTM$^\ast$\cite{yao2018exploring}&80.9&38.3 &28.6 &58.5 &128.7 &22.1\\
    AUTO-ENC\cite{yang2019auto}&80.8  &38.4 & 28.4 & 58.6 & 127.8 &22.1\\
    ALV$^\ast$\cite{guo2019aligning}&-  &38.4 &28.5 &58.4 & 128.6 & 22.0\\
    GCN-LSTM-HIP$^{\ast\dagger}$\cite{yao2019hierarchy}&-&39.1 &28.9 &\textbf{59.2} &130.6 &22.3\\
    \hline
    \multicolumn{7}{c}{\textbf{transformer based model}}\\
    \hline
    Entangle-T$^\ast$\cite{li2019entangled}&\textbf{81.5}& \textbf{39.9} & 28.9 &59.0 & 127.6 &22.6\\
    AoA\cite{huang2019attention}&80.2 &38.9 &\textbf{29.2} &58.8 & 129.8 &22.4\\
    VORN\cite{herdade2019image}&80.5 & 38.6 & 28.7 & 58.4 & 128.3& 22.6\\
    Ours&80.8 & 39.5 & 29.1 & 59.0 & \textbf{130.8}& \textbf{22.8}\\
    \hline
    \end{tabular}
    \caption{Comparison on MSCOCO Karpathy offline test split. $^\ast$ means fusion of two models. $^\dagger$ means SENet \cite{hu2018squeeze} as feature extraction backbone}
    \label{tab:my_label}
\vspace{-25pt}
\end{table}
\section{Experiment}\label{sec:experiments}
\subsection{Datasets and Evaluation Metrics}
Our model is trained on the MSCOCO image captioning dataset~\cite{chen2015microsoft}. We follow Karpathy's splits \cite{karpathy2015deep}, with 11,3287 images in the training set, 5,000 images in the validation set and 5,000 images in the test set. Each image has 5 captions as ground truth. We discard the words which occur less than 4 times, and the final vocabulary size is 10,369. We test our model on both Karpathy's offline test set (5,000 images) and MSCOCO online testing datasets (40,775 images). We use Bleu \cite{papineni2002bleu}, METEOR \cite{banerjee2005meteor}, ROUGE-L \cite{lin2004rouge}, CIDEr \cite{vedantam2015cider}, and SPICE \cite{anderson2016spice} as evaluation metrics.

\subsection{Implementation Details}
Following previous work, we first train Faster R-CNN on Visual Genome \cite{krishna2017visual}, use resnet-101 \cite{he2016deep} as backbone, pretrained on ImageNet \cite{deng2009imagenet}. For each image, we can detect $10-100$ informative regions, the boundaries of each are first normalised and then used to compute the spatial graph matrices. We then train our proposed model for image captioning using the computed spatial graph matrices and extracted features for each image region. We first train our model with \emph{cross-entropy} loss for 25 epochs, the initial learning rate is set to $2\times 10^{-3}$, and we decay the learning rate by $0.8$ every 3 epochs. Our model is optimized through Adam \cite{kingma2014adam} with a batch size of 10. We then further optimize our model by reinforced learning for another 35 epochs. The size of the decoder's LSTM layer is set to 1024, and beam search of size 3 is used in the inference stage.

\subsection{Experiment Results}
We compare our model's performance with published image captioning models. The compared models include the top performing single-stage attention model, Att2all \cite{rennie2017self}; two-stages attention based models, n-babytalk \cite{lu2018neural} and up-down \cite{anderson2018bottom}; visual scene graph based models, GCN-LSTM \cite{yao2018exploring}, AUTO-ENC \cite{yang2019auto}, ALV \cite{guo2019aligning}, GCN-LSTM-HIP \cite{yao2019hierarchy}; and transformer based models Entangle-T \cite{li2019entangled}, AoA \cite{huang2019attention}, VORN \cite{herdade2019image}. The comparison on the MSCOCO Karpathy offline test set is illustrated in Table~\ref{tab:my_label}. Our model achieves new state-of-the-art on the CIDEr and SPICE score, while other evaluation scores are comparable to the previous top performing models. Note that because most visual scene graph based models fused semantic and spatial scene graph, and require the auxiliary models to build the scene graph at first, our model is more computationally efficient. VORN \cite{herdade2019image} also integrated spatial attention in their model, and our model performs better than them among all kinds of evaluation metrics, which shows the superiority of our spatial graph transformer layer.
The MSCOCO online testing results are listed in Tab.~\ref{tab:online_test}, our model outperforms previous transformer based model on several evaluation metrics.

\begin{table}[t!]
    \centering
    \begin{tabular}{c|c c| c c| c c| c c| c c}
    \hline
    \multirow{2}{*}{model} & \multicolumn{2}{c }{B1} & \multicolumn{2}{c }{B4} & \multicolumn{2}{c }{M} & \multicolumn{2}{c }{R} & \multicolumn{2}{c }{C}\\
    \cline{2-11}
    &c5&c40&c5&c40&c5&c40&c5&c40&c5&c40\\
    \hline
    \multicolumn{11}{c}{\textbf{scene graph based model}}\\
    \hline
    GCN-LSTM$^\ast$\cite{yao2018exploring}&80.8&95.9&38.7&69.7&28.5&37.6&58.5&73.4&125.3&126.5\\
    \hline
    AUTO-ENC$^\ast$\cite{yang2019auto}& -&-&38.5&69.7&28.2&37.2&58.6&73.6&123.8&126.5\\
    \hline
    ALV$^\ast$\cite{guo2019aligning}&79.9&94.7&37.4&68.3&28.2&37.1&57.9&72.8&123.1&125.5\\
    \hline
    GCN-LSTM-HIP$^{\ast\dagger}$\cite{yao2019hierarchy}&81.6&95.9&39.3&71.0&28.8&38.1&59.0&74.1&127.9&130.2\\
    \hline
    \multicolumn{11}{c}{\textbf{transformer based model}}\\
    \hline
    Entangle-T$^\ast$\cite{li2019entangled}&81.2&95.0&38.9&70.2&28.6&38.0&58.6&73.9&122.1&124.4\\
    \hline
    AoA\cite{huang2019attention}&81.0&95.0&39.4&71.2&29.1&38.5&58.9&74.5&126.9&129.6\\
    \hline
    Ours&81.2&95.4&39.6&71.5&29.1&38.4&59.2&74.5&127.4&129.6 \\
    \hline
    \end{tabular}
    \caption{Leaderboard of recent published models on the MSCOCO online testing server. $^\ast$ means fusion of two models. $^\dagger$ means SENet \cite{hu2018squeeze} as feature extraction backbone}
    \label{tab:online_test}
\vspace{-25pt}
\end{table}

\subsection{Ablation Study and Analysis}

In the ablation study, we use AoA~\cite{huang2019attention} as a strong baseline \footnote{\scriptsize Our experiments are based on the code released at: \url{https://github.com/husthuaan/AoANet}} (with a single multi-head dot-product attention module per layer), which add the gated linear layer \cite{dauphin2017language} on top of the multi-head attention. In the encoder part, we study the spatial relationship's effect in the encoder, where we ablate the spatial relationship by simply taking the mean output of three sub-transformers in each layer by reformulating Eqs.~\ref{eq:6} and~\ref{eq:7} as: $\text{Attention}(Q,K_i,V_i) = \text{Softmax}\left(\frac{QK_{i}^{T}}{\sqrt{d}}\right)V_{i},A_m = \text{Norm}\left(A+\frac{1}{3}\sum_{i \in \{p,n,c\}}\text{MultiHead}(Q,K_i,V_i)\right)$. We also study where to use our proposed spatial graph encoding transformer layer in the encoding part: in the first layer, second layer, third layer or three of them? In the decoding part, we study the effect of the  number of sub-transformers ($M$ in Eq.~\ref{eq:eq10}) in the implicit decoding transformer layer.

\begin{table}[h!]
\begin{minipage}{\textwidth}  
\begin{center}
    \begin{tabular}{c|c c c c c c}
    \hline
    model & Bleu1 & Bleu4 & METEOR & ROUGE-L & CIDEr & SPICE\\
    \hline
    baseline(AoA) & 77.0 & 36.5 & 28.1 & 57.1 & 116.6 & 21.3 \\
    \hline
    \multicolumn{7}{c}{\textbf{positions to embed our spatial graph encoding transformer layer}}\\
    \hline
    baseline+layer1     &77.8 &36.8 & 28.3 & 57.3 & 118.1 & 21.3\\
    baseline+layer2     & 77.2 &36.8 & 28.3 & 57.3 & 118.2 & 21.3\\
    baseline+layer3     &  77.0 &37.0 & 28.2 & 57.1 & 117.3 & 21.2\\
    baseline+layer1,2,3     &  77.5 &37.0 & 28.3 & 57.2 & 118.2 & 21.4\\
    \hline
    \multicolumn{7}{c}{\textbf{effect of spatial relationships in the encoder}}\\
    \hline
    baseline+layer1,2,3 w/o spatial rela     &  77.5 &36.8 & 28.2 & 57.1 & 117.8 &21.4\\
    \hline
    \multicolumn{7}{c}{\textbf{number of sub-transformers in the implicit decoding transformer layer}}\\
    \hline
    baseline+layer1,2,3 (M=2)     &77.5   &37.6 &28.4  &57.4  &118.8  &21.3 \\
    baseline+layer1,2,3 (M=3)  &78.0&  37.4 &28.4 & 57.6 & 119.1 &  21.6 \\
    baseline+layer1,2,3 (M=4)     &77.5   &37.8 &28.4  &57.5  &118.6  &21.4 \\
    \hline
    \end{tabular}
    \caption [caption]{Ablation study, results reported without RL training. baseline+layer1 means only the first layer of encoding transformer uses our proposed spatial transformer layer, other layers use the original one. $M$ is the number of sub-transformers in the decoding transformer layer.}
    \label{tab:ab_study}
    \end{center}
    \end{minipage}
\vspace{-25pt}
\end{table}

\begin{figure}[t!]
    \centering
    \includegraphics[width=1.0\linewidth]{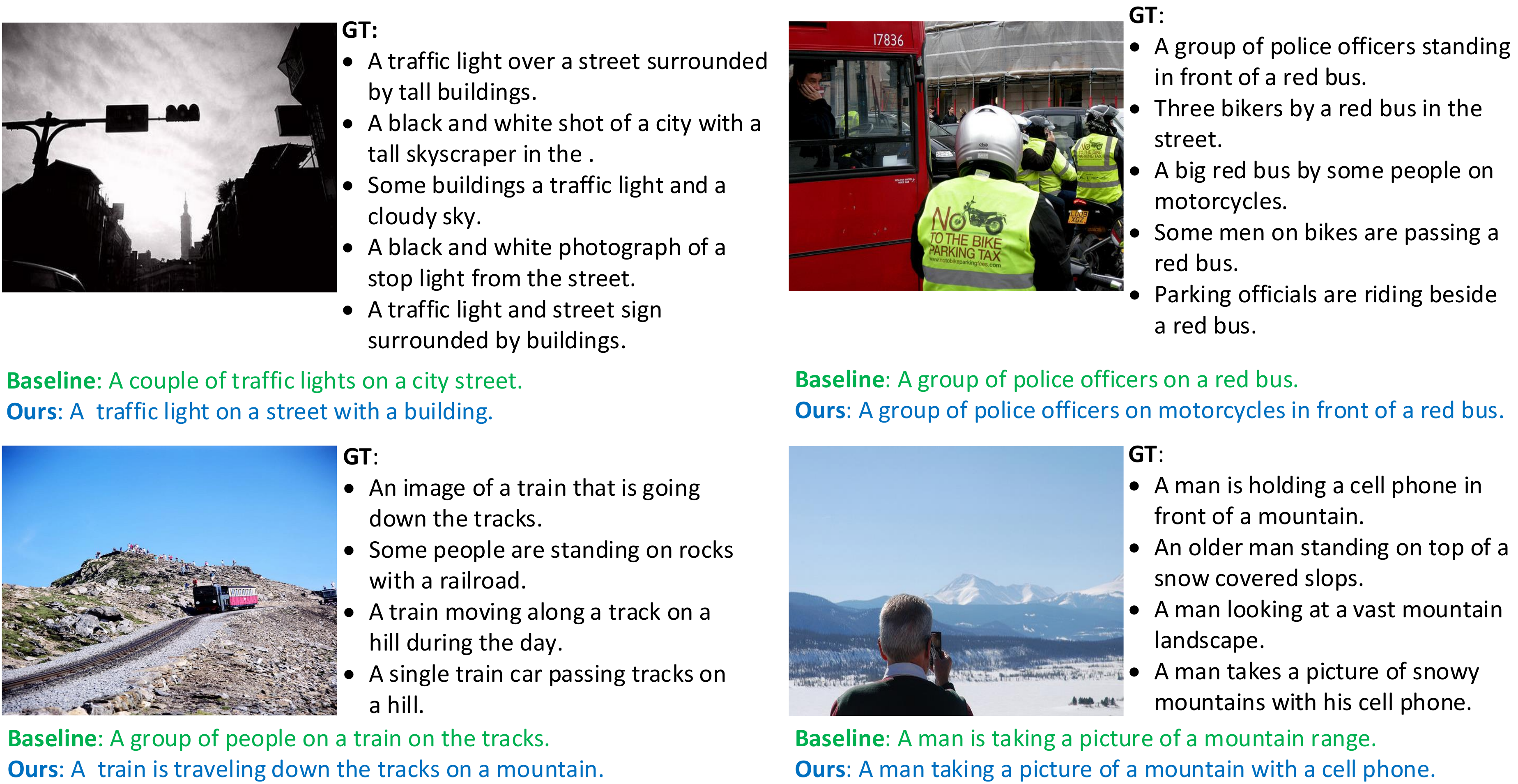}
    \caption{Qualitative examples from our method on the MSCOCO image captioning dataset~\cite{chen2015microsoft}, compared against the ground truth annotation and a strong baseline method (AoA \cite{huang2019attention}).}
    \label{fig:qualitative}
\vspace{-15pt}
\end{figure}

As we can see from Tab.~\ref{tab:ab_study}, by widening the encoding transformer layer, there is a significant improvement on the model's performance. While not every layers in the encoding transformer are equal, when we use our proposed transformer layer at the top layer of the encoding part, the improvement was reduced. This may be because spatial relationships at the top layer of the transformer are not as informative, we use our spatial transformer layer at all layers in the encoding part. When we reduce the spatial relationship in our proposed wider transformer layer, there is also some performance reduction, which shows the importance of the spatial relationship in our design. After widening the decoding transformer, the improvement was further increased (the CIDEr score increased from 118.2 to 119.1 after widening the decoding transformer layer with 3 sub-transformers), while not more wider gives better result, with 4 sub-transformers in the decoding transformer layer, there is some performance decrease, therefore the final design of our decoding transformer layer has 3 sub-transformers in parallel. The qualitative example of our models results is illustrated in  Fig.~\ref{fig:qualitative}. As we can see, the baseline model without spatial relationships wrongly described the police officers on a red bus (top right), and people on a train (bottom left).

%\footnotetext{Our experiment is based the code released at: \url{https://github.com/husthuaan/AoANet}}

\begin{figure}[t!]
    \centering
    \includegraphics[width=1.0\linewidth]{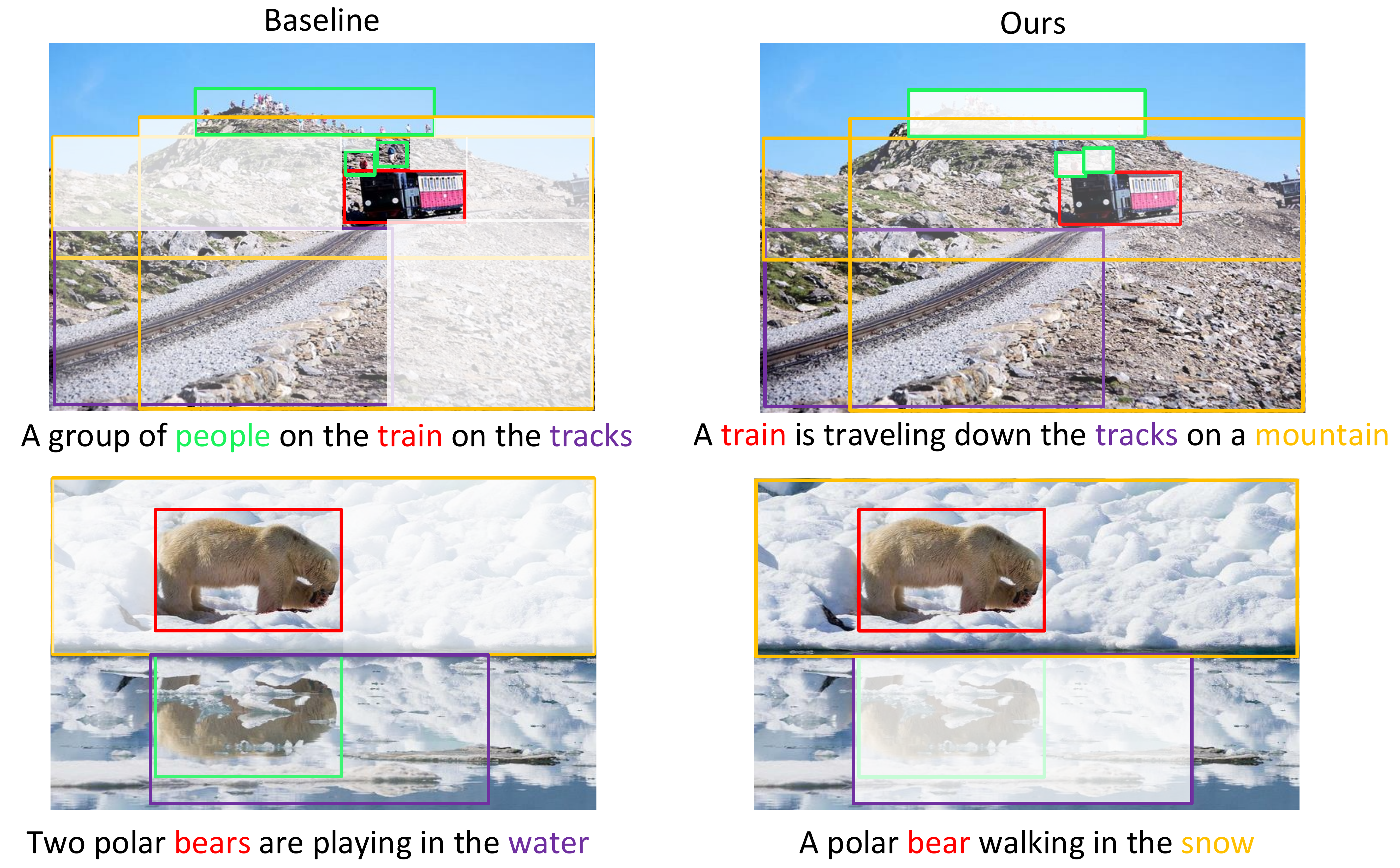}
    \caption{A visualization of how the query region relates to its other key regions through attention, the region in the red bounding box is the query region and other regions are key regions. The transparency of each key region shows its dot-product attention weight with the query region. Higher transparency means larger dot-product attention weight, vice versa.}
    \label{fig:attetion1}
    \centering
\vspace{-15pt}
\end{figure}

\paragraph{Encoding implicit graph visualisation:} the transformer layer can be seen as an implicit graph, which relates the informative regions through dot-product attention. Here we visualise how our proposed spatial graph transformer layer learn to connect the informative regions through attention in Fig.~\ref{fig:attetion1}. In the top example, the original transformer layer strongly relates the train with the people on the mountain, yields wrong description, while our proposed transformer layer relates the train with the tracks and mountain; in the bottom example, the original transformer relates the bear with its reflection in water and treats them as `two bears', while our transformer can distinguish the bear from its reflection and relate it to the snow area. 

\begin{figure}[t!]
\centering
\begin{subfigure}[b]{0.45\linewidth}
\includegraphics[width=\linewidth]{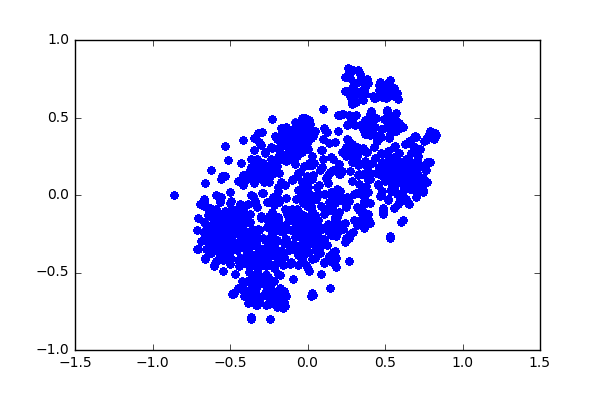}
\caption{original}
\end{subfigure}
\begin{subfigure}[b]{0.45\linewidth}
\includegraphics[width=\linewidth]{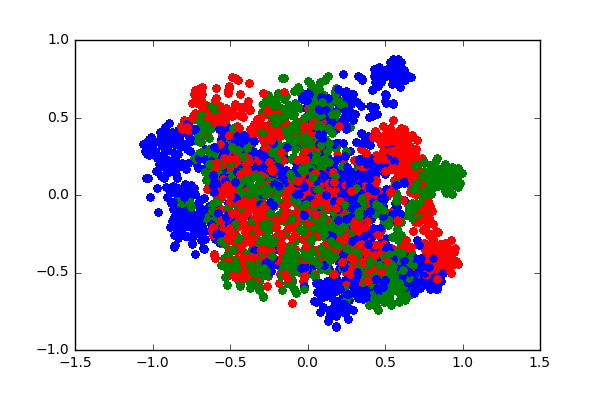}
\caption{ours}
\end{subfigure}
%\vspace{-10pt}
\caption{t-SNE \cite{maaten2008visualizing} visualisation of the output from decoding transformer layer (1,000 examples), different color represent the output from different sub-transformers in the decoder in our model.}
\label{fig:decoder_out}
\vspace{-15pt}
\end{figure}

\paragraph{Decoding feature space visualisation:} We also visualised the output of our decoding transformer layer (Fig.~\ref{fig:decoder_out}). Compared to the original decoding transformer layer, which only has one sub-transformer inside it. The output of our proposed implicit decoding transformer layer covers a larger area in the reduced feature space than the original one, which means that our decoding transformer layer decoding more information in the image regions. In the original feature space (1,024 dimensions) from the output of decoding transformer layer, we compute the trace of the feature maps' co-variance matrix from 1,000 examples, the trace for original transformer layer is $30.40$ compared to $454.57$ for our wider decoding transformer layer, which indicates that our design enables the decoder's output to cover a larger area in the feature space. However, it looks like individual sub-transformers in the decoding transformer layer still do not learn to disentangle different factors in the feature space (as there is no distinct cluster from the output of each sub-transformer), we speculate this is because we have no direct supervision to their output, which may not able to learn the disentangled feature automatically \cite{locatello2019challenging}.

\section{Discussion and Conclusion}
\label{sec:conclusion}

In this work, we introduced the \textbf{\textit{image transformer}} architecture. The core idea behind the proposed architecture is to widen the original transformer layer, designed for machine translation, to adapt it to the structure of images. In the encoder, we widen the transformer layer by exploiting the spatial relationships between image regions, and in the decoder, the wider transformer layer can decode more information in the image regions. Extensive experiments were done to show the superiority of the proposed model, the qualitative and quantitative analyses were illustrated in the experiments to validate the proposed encoding and decoding transformer layer. Compared to the previous top models in image captioning, our model achieves a new state-of-the-art SPICE score, while in the other evaluation metrics, our model is either comparable or outperforms the previous best models, with a better computational efficiency.

 We hope our work can inspire the community to develop more advanced transformer based architectures that can not only benefit image captioning but also other computer vision tasks which need relational attention inside it. Our code will be shared with the community to support future research.

%===========================================================
\bibliographystyle{splncs}
\bibliography{egbib}

%this would normally be the end of your paper, but you may also have an appendix
%within the given limit of number of pages
\end{document}